\documentclass{article}

\usepackage[preprint]{neurips_2024}

\usepackage[utf8]{inputenc} 
\usepackage[T1]{fontenc}    
\usepackage{hyperref}       
\usepackage{url}            
\usepackage{booktabs}       
\usepackage{amsfonts}       
\usepackage{nicefrac}       
\usepackage{microtype}      
\usepackage{xcolor}         
\usepackage{graphicx}
\usepackage{mdframed}
\usepackage{listings}
\usepackage[most]{tcolorbox}
\tcbuselibrary{listingsutf8}
\usepackage{caption}      
\usepackage{subcaption} 

\usepackage{pifont}   

\newcommand{\cmark}{\textcolor{green}{\ding{51}}}  
\newcommand{\xmark}{\textcolor{red}{\ding{55}}}    

\usepackage[normalem]{ulem}

\newcommand{\apigen}{APIGen~\citep{chen2024apigen}}
\newcommand{\name}{SEAL}
\usepackage{ifthen}

\newboolean{draft}
\setboolean{draft}{False}

\ifthenelse{\boolean{draft}}{
    \newcommand{\sasha}[1]{\textcolor{green}{Sasha: #1}}
    \newcommand{\woojeong}[1]{\textcolor{blue}{Woojeong: #1}}
    \newcommand{\ashish}[1]{\textcolor{purple}{Ashish: #1}}
    \newcommand{\aditya}[1]{\textcolor{pink}{Aditya: #1}}
}{
    \newcommand{\sasha}[1]{}
    \newcommand{\woojeong}[1]{}
    \newcommand{\ashish}[1]{}
    \newcommand{\aditya}[1]{}
}

\title{\name: Suite for Evaluating API-use of LLMs}

\author{%
  Woojeong Kim \\
  Cornell University\\
  \texttt{wk247@cornell.edu}
  \And
  Ashish Jagmohan \\
  Emergence AI \\
  \texttt{ashish@emergence.ai} \\
  \And
  Aditya Vempaty \\
  Emergence AI \\
  \texttt{aditya@emergence.ai} \\
}

\begin{document}

\maketitle

\begin{abstract}

Large language models (LLMs) have limitations in handling tasks that require real-time access to external APIs. 
While several benchmarks like ToolBench and APIGen have been developed to assess LLMs' API-use capabilities, they often suffer from issues such as lack of generalizability, limited multi-step reasoning coverage, and instability due to real-time API fluctuations.
In this paper, we introduce \name, an end-to-end testbed designed to evaluate LLMs in real-world API usage. \name~standardizes existing benchmarks, integrates an agent system for testing API retrieval and planning, and addresses the instability of real-time APIs by introducing a GPT-4-powered API simulator with caching for deterministic evaluations. Our testbed provides a comprehensive evaluation pipeline that covers API retrieval, API calls, and final responses, offering a reliable framework for structured performance comparison in diverse real-world scenarios.
\name~is publicly available, with ongoing updates for new benchmarks.


\end{abstract}

\section{Introduction}
While large language models (LLMs) excel at many language tasks, they face limitations when handling tasks that require real-time access to specific information, such as current events, calculations, or web searches. Tools like calculators, code execution, and browsing extend LLMs’ capabilities, enabling them to perform specialized tasks and access up-to-date knowledge, adapting dynamically to users’ needs. Recently, several benchmarks have been introduced to assess LLMs’ ability to interact with real-world APIs, moving beyond a small set of hand-coded tools to a broader pool of practical, real-world applications. These include ToolBench~\citep{qin2023toolllm} and variants thereof, APIGen~\citep{chen2024apigen}, AnyTool~\citep{du2024anytool} and MetaTool~\citep{huang2023metatool}. 

In this paper, we analyze major API-use\footnote{We use ``tool'' and ``API'' interchangeably throughout this paper. While we primarily focus on APIs, the same methodology can be extended to general tool usage with appropriate API-wrapper around them.} benchmarks and argue that there are critical gaps. Specifically, we identify a number of common issues, including the lack of clear holdout sets leading to overfitting, poor coverage of multi-step reasoning queries which are essential in real-world use cases of such systems, and deficiencies in benchmark quality and stability. Further, some benchmarks only focus on certain aspects; For example, AnyTool~\citep{du2024anytool} and MetaTool~\citep{huang2023metatool} focus on tool selection but overlook other important aspects such as the content of tool calls and the final response. Similarly, \apigen~tests function-calling capabilities but neglects multiple possible trajectories with only one hard-coded answer.



In response, we present \name, a comprehensive, end-to-end testbed for evaluating LLMs in tool usage, particularly with diverse real-world APIs. This testbed provides uniform test environment of API calling system by sanitizing and standardizing existing benchmarks, incorporating an agent system built on AutoGen~\citep{wu2023autogen} for testing both API retrieval and planning, and providing a robust evaluation pipeline. 
Due to the real-time nature of APIs, existing benchmarks often lack reliable evaluation as there are no static ground truth answers.
To address this, we developed an API simulator powered by GPT-4~\citep{achiam2023gpt} to generate plausible API responses, building on the approach in StableToolbench~\citep{guo2024stabletoolbench}. 
We further enhance this system with caching to enable more deterministic evaluations. 
Additionally, we offer a comprehensive evaluation framework that covers all aspects of API learning, including API retrieval, API calls, and the final response.
Our testbed allows users to test their retrieval and planning methods while providing a more structured and reliable performance comparison.


Our contributions are as follows:
\begin{itemize}
    \item We investigate the shortcomings of existing API-use benchmarks, identifying issues like lack of generalizability, limited multi-step reasoning coverage, and benchmark instability.
    \item We present \name, a comprehensive testbed that standardizes existing benchmarks and incorporates an agent-based system for evaluating LLMs in retrieval, planning, and execution.
    \item We offer an end-to-end evaluation framework covering API retrieval, API calls, and final responses, enabling structured performance comparison across diverse real-world scenarios.
\end{itemize}


\begin{figure}[!t]
    \centering
    \includegraphics[width=0.8\textwidth]{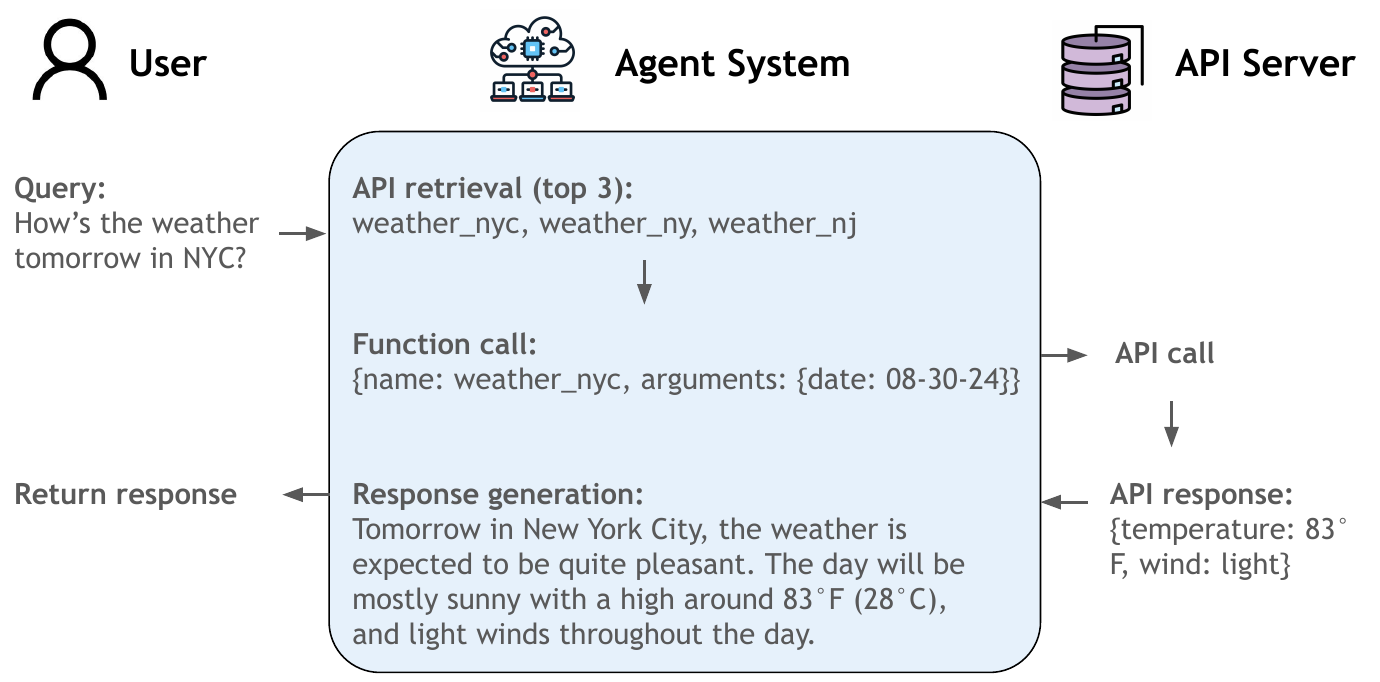}
    \caption{Workflow of a single-step, single-API-use system}
    \label{fig:workflow}
\end{figure}

\begin{table}[!t]
\caption{Comparison of tool-use benchmarks. For datasets with multiple subsets, we only report statistics for the largest subset: G1-train for Toolbench, and Huggingface for APIBench}
\label{table:benchmark_comparison}
\centering
\resizebox{\textwidth}{!}{
\begin{tabular}{@{}llllll@{}}
\toprule
 & \begin{tabular}[c]{@{}l@{}}Toolbench \\ \cite{qin2023toolllm}\end{tabular} & \begin{tabular}[c]{@{}l@{}}APIGen \\ \citep{chen2024apigen} \end{tabular} & \begin{tabular}[c]{@{}l@{}}AnyTool \\ \citep{du2024anytool} \end{tabular} & \begin{tabular}[c]{@{}l@{}}MetaTool \\ \citep{huang2023metatool} \end{tabular} &
 \begin{tabular}[c]{@{}l@{}}APIBench \\ \cite{patil2023gorilla}\end{tabular} \\
 \midrule
API source & RapidAPI & RapidAPI & RapidAPI & OpenAI plugin & HuggingFace \\
Total \# of queries & 40399 & 60000 & 407 & 21047 & 8191 \\
- multi-step only & 34052 & 20448 & 339 & 497 & 0 \\
Avg. \# of APIs per query & 4.5 & 1.7 & 2.3 & 1.1 & 1.0 \\ \midrule
\# of APIs with queries & 8684 & 3179 & 307 & 199 & 914 \\
Total \# of APIs & 16464 & 3605 & 16464 & 437 & 914 
\end{tabular}
}
\end{table}

\section{Overall Landscape}
\label{sec:workflow}
Fig.~\ref{fig:workflow} provides an overview of the typical workflow for tool-use systems. These systems consist of three main components: the user, the agent system, and the API server. The process begins when a user queries the agent system, for example, ``How’s the weather tomorrow in NYC?'' The next step typically is a retrieval to identify relevant APIs for the language model that powers the agentic system. This is especially useful in scenarios with numerous real-world APIs, where providing all available APIs as input is impractical due to context limitations (such as length, and lost-in-the-middle phenomenon). The LLM then generates the arguments for tool calls based on the API documentation or the tool specifications and forwards them to the API server. Typically, the API server operates as a separate component outside the agent system. Once the API response is received, the agent system's final role is to generate a response summarizing the API’s output. While this example involves a single-step process, in many real-world cases, these steps are repeated, with each step’s input determined by the output of the previous step. There may also be an optional verification step to ensure intermediate and final responses are generated correctly.

This is a multi-step, complex process that relies on several LLM capabilities, including accurate retrieval, correct function calling, and factual summarization. However, existing benchmarks focus only on specific aspects of this pipeline: AnyTool and MetaTool emphasize tool selection, while APIGen focuses on function-calling capabilities. Section~\ref{sec:challenges} talks about these challenges in more detail.

Another often-overlooked aspect is that this process involves collaboration between multiple agents, such as the tool retriever, the main orchestrator, and optionally, a planner or verifier.~\citep{mekala2024toolverifier, shen2024small}
Most previous benchmarks~\citep{chen2024apigen, qin2023toolllm, patil2023gorilla} customize a single LLM to handle all these steps. We argue that this could be tackled more effectively using an agent system, where certain components are powered by an LLM. In fact, this separation is already happening internally, though it’s not often recognized as distinct agent components. For example, tool retrieval is often handled independently of the LLM, using query/API embeddings generated by large embedding models or smaller sentence-transformer models, followed by vector lookups. Our testbed integrates these components under the concept of agents, and we implement it using the AutoGen~\citep{wu2023autogen} framework. Section~\ref{sec:testbed} discusses this in more detail.

\section{Challenges in Existing Benchmarks}
\label{sec:challenges}

\begin{figure}[!t]
    \centering
    \includegraphics[width=\textwidth]{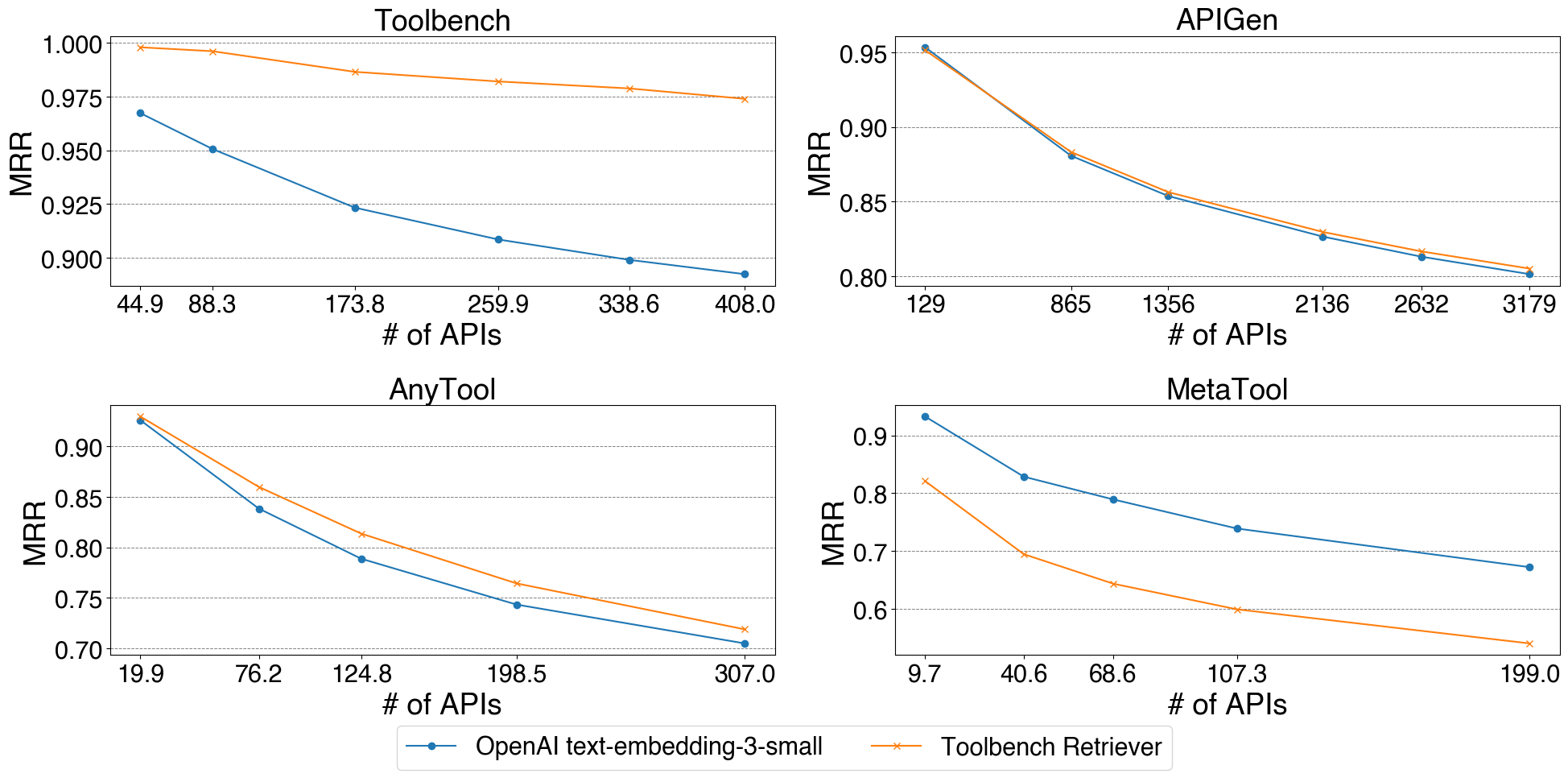}
    \vspace{-0.1in}
    \caption{Comparison of embedding models' API retrieval performance across benchmarks. We fixed the total number of sampled queries and report the average performance over 10 sampling runs. For ToolBench, results are based on the test split, as the ToolBench retriever was trained on the train split.}
    \vspace{-0.2in}
    \label{fig:retrieval}
\end{figure}


We conduct an in-depth evaluation of existing benchmarks and identified several critical limitations that restrict their wider applicability. Table~\ref{table:benchmark_comparison} presents key statistics for five benchmarks we focus on. While these are the primary benchmarks under discussion, there are additional ones within the scope that we plan to further support. Our analysis highlights the following four significant limitations.

\subsection{Lack of Generalizability}
Previous works have claimed that small, open-source models trained on specific benchmarks can perform as well as, or better than, general-purpose models. However, as noted by \citet{kapoor2024ai}, many agent benchmarks lack adequate holdout sets, or even fail to include them entirely. This is also true for API-calling benchmarks. For instance, APIGen does not provide a clear train-test split, making it difficult to evaluate models using the provided benchmark data. Similarly, ToolBench includes a very small test set that fails to represent the diversity of real-world scenarios. This limitation in benchmark design often results in models that perform well on narrow datasets but struggle with generalization. We illustrate this through an example of API retrievers, as retrieving the appropriate APIs is critical to overall performance. Previous works~\citep{qin2023toolllm, kong2023tptu} trained small sentence transformer models~\citep{reimers2019sentence} using contrastive learning objectives, and claim superior performance over proprietary, general-purpose embedding models like OpenAI’s. However, we find that while these trained retrievers performed well on the same source of APIs, their performance significantly declines when applied to different sources. 
As shown in Figure~\ref{fig:retrieval}, the retriever trained on ToolBench outperforms OpenAI’s embedding model on three RapidAPI-based benchmarks but performs significantly worse on MetaTool, a dataset based on OpenAI plugins. This highlights the importance of generalization to various API sets, a critical feature in realistic scenarios where APIs are frequently updated, deprecated, or custom-built by users. 

\subsection{Bias Towards Simple Queries}

To better simulate realistic scenarios, we emphasize the need for multi-tool and multi-step benchmarks. As state-of-the-art LLMs become increasingly adept at basic function calls~\citep{qu2024tool}, it is crucial to assess their ability to decompose complex user queries into smaller, executable substeps and plan actions accordingly. Unfortunately, most existing benchmarks consist predominantly of single-step queries, where a single API call is sufficient to complete the task, as shown in Fig~\ref{table:benchmark_comparison}. Notably, one of the most widely-used benchmarks, APIBench~\citep{patil2023gorilla}, consists solely of single-step and single-tool queries. Existing benchmarks also lack realistic queries that require sequentially dependent reasoning. We show a few examples in the Appendix~\ref{app:easy_query}.


\subsection{General Instability}

A significant issue with current benchmarks is their instability, as static ground truth quickly becomes outdated, and API services exhibit variability. API responses are time-sensitive, and services change over time due to factors like deprecation, shifts in service definitions, and altered response behaviors. This instability makes it difficult to evaluate new systems on older benchmarks, hindering efforts to standardize evaluation. To mitigate these challenges, several benchmarks restrict themselves to small sets of hard-coded and deterministic tools~\citep{schick2024toolformer, lu2024toolsandbox}, instead of dynamic real-world APIs. Other benchmarks limit themselves in both API pool size and the number of queries.




One widely used benchmark for API-use, that does not limit its scope as above, is the ToolBench benchmark. Despite the efforts of ToolBench to enhance stability by introducing a proxy server for RapidAPI, which simplifies access by handling authentication and overhead associated with real-time APIs, we found ToolBench to be highly unstable. Many APIs fail to return consistent responses and frequently produce a variety of errors. Previous work~\citep{guo2024stabletoolbench} highlighted ToolBench's instability and proposed solutions like caching API responses and using a GPT-4-based simulator for unresponsive APIs. However, these fixes were limited to a small subset of ToolBench, and running LLM simulators on demand still introduces stochastic variability. 



\subsection{Incomplete Evaluation}
\label{sec:challenges:eval}

\begin{table}[]
\caption{Comparison of benchmarks on evaluation}
\label{table:evaluation}
\centering
\begin{tabular}{@{}llllll@{}}
\toprule
 & \textbf{Toolbench} & \textbf{APIGen} & \textbf{AnyTool} & \textbf{MetaTool} & \textbf{APIBench} \\ \midrule
API retrieval & \cmark & \cmark & \cmark & \cmark & \xmark \\
API call & \xmark & \cmark & \xmark & \xmark & \xmark \\
Final response & \cmark & \xmark & \xmark & \xmark & \cmark \\ \bottomrule
\end{tabular}
\vspace{-0.2in}
\end{table}

For a comprehensive evaluation of API-calling systems, it is essential to assess each stage of the pipeline: whether the correct tools are retrieved, whether the correct tools are called, whether the tool calls are accurate, and finally, whether the query response is correct. The real-time nature of APIs further complicates the evaluation of API-use systems, and existing works focus only on partial components of the overall pipeline described in Section~\ref{sec:workflow}. 
As shown in Table~\ref{table:evaluation}, AnyTool and MetaTool only evaluate API retrieval, while APIGen evaluates both API retrieval and calls but overlooks the fact that multiple trajectories can lead to a successful final response. Although ToolBench aims to provide ground truth across the entire pipeline, the provided ground truth labels for tool calls and final responses is somewhat outdated and unreliable. Approximately 40\% of queries end up as "unsolvable," where the LLM used in ToolBench's own implementation fails to generate a valid output. Each stage of the API-calling system needs to be evaluated to enable more reliable and comprehensive assessment.

\section{\name~Construction}
\label{sec:testbed}
\name is a comprehensive, end-to-end testbed that builds on top of current benchmarks. It includes standardization and sanitization of queries from existing benchmarks, a flexible agent system capable of adapting to user demands, and a thorough evaluation pipeline.

\subsection{Benchmark Standardization \& Sanitization}

We parse and standardize five existing benchmarks introduced in Table~\ref{table:benchmark_comparison}, into a unified format consisting of queries, APIs, query-to-API mapping, and query-to-API-call mapping. Note that API-call data is available for only 3 out of the 5 datasets, so this field remains empty for AnyTool and MetaTool. This standardized format enables the use of multiple benchmarks with diverse structures via a unified approach. For instance, users can search for APIs associated with a specific query ID or retrieve an API by its name or features. Data sanitization and filtering details are in the Appendix~\ref{app:standardization}.

\subsection{Agent System Construction}

\begin{figure}[!t]
    \centering
    \includegraphics[width=0.9\textwidth]{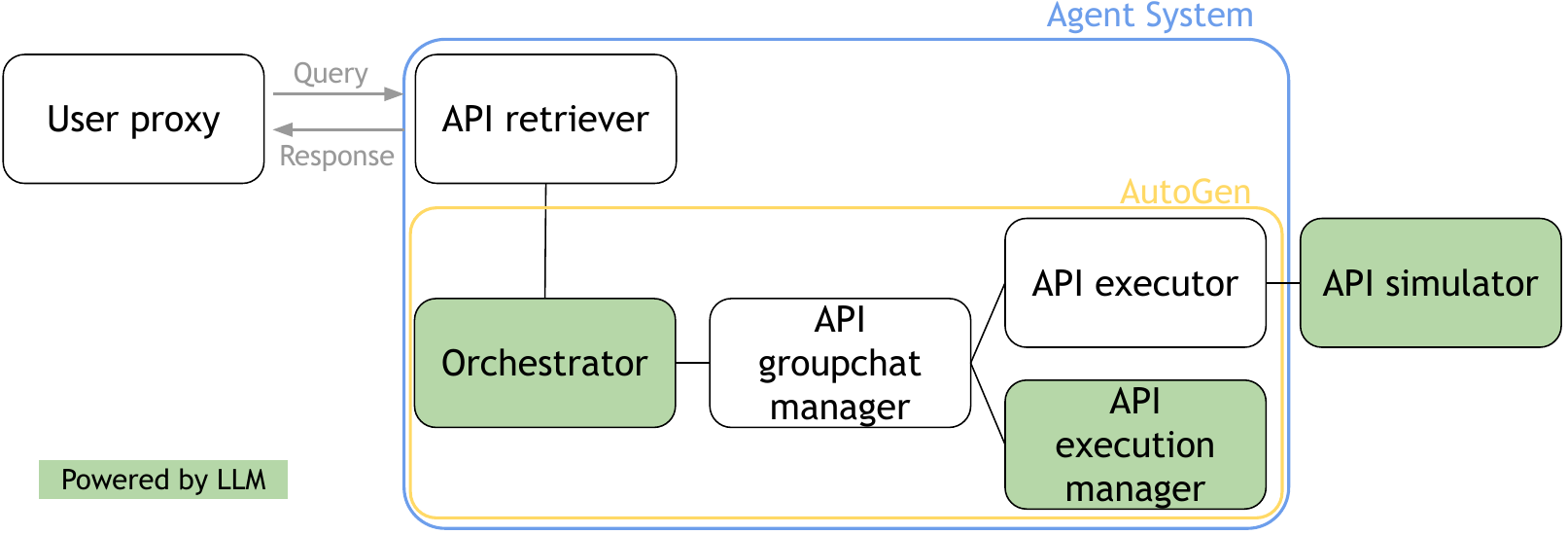}
    \caption{AutoGen system architecture}
    \label{fig:autogen_architecture}
\end{figure}

Previous works have developed single LLMs to handle every step of the API-calling process. However, we argue that API-calling, and tool-calling in general, can be tackled more effectively using an agent-based system, where multiple agents collaborate, with some powered by LLMs. This design of separation allows for testing both tool retrieval and tool planning methods, enabling users to easily swap different agents in and out. It also opens up the possibility of replacing agents with smaller, specialized models, rather than relying on expensive general models to handle everything. We develop an adaptable agent system based on the AutoGen~\citep{wu2023autogen} framework. One of the key advantages of AutoGen is its customizable agents, which can operate in various modes by leveraging combinations of LLMs, human inputs, and tools/APIs. This flexibility allows the system to dynamically adjust agent behaviors based on user-specified configurations.

Fig.~\ref{fig:autogen_architecture} illustrates the current architecture of \name. We follow the workflow of existing systems, where APIs are pre-selected and registered within the agent system. Within this system, API-calling is managed through a ``group chat'' interaction between a API Executor and a API Execution Manager. The API Execution Manager, powered by an LLM, generates the necessary API calls and arguments, while the API Executor communicates with a API Simulator in the background to simulate real API servers. Although this represents the simplest version of the architecture, the system is designed to allow easy addition or removal of agents. For instance, one could add a Planner Agent under the Orchestrator to break down multi-step queries into smaller sub-steps or introduce a Verifier Agent to ensure that each step has been executed correctly.

\paragraph{API Retriever}
Given the large size of the API pool in API-calling benchmarks, it is impractical to register all APIs with an LLM due to context length limits. We implement an API retriever as a separate class, allowing users to specify the embedding model of their choice. Available options include OpenAI embedding models, Gemini embedding models, and a customized sentence transformer model proposed by \cite{qin2023toolllm}. Once the user query and API documents are embedded, the retriever efficiently performs vector lookups using the Faiss library.

\paragraph{API Simulator}

Since real APIs are inherently real-time, and we find existing benchmarks' API servers to be highly unstable, we develop a custom API simulator powered by GPT-4. Similar approach is used in \cite{guo2024stabletoolbench}, where real-time API responses are cached and the simulator is called on demand. In our case, we fully replace and cache all API responses with simulations to minimize stochastic variability. Given instructions for simulation, API documentation, and API parameters, the simulator replicates API behavior. The simulator prompt can be found in Appendix~\ref{app:simalutor_prompt}. Although we acknowledge that LLM-based systems may generate fabricated information, we believe it is essential to have reliable and deterministic APIs from the perspective of benchmarking.

\subsection{Evaluation Pipeline}
Previous works evaluate one of the following: (1) the final response of the system~\citep{qin2023toolllm}, considering that multiple reasoning trajectories can lead to success, or (2) a single hard-coded trajectory by checking if the API calls are correct~\citep{chen2024apigen}. We argue that a more comprehensive evaluation is needed, spanning the entire pipeline of API usage, including API retrieval, API calls, and the final response.

\begin{itemize} 
    \item \textbf{API Retrieval:} Did the system retrieve the correct tools? All five benchmarks listed in Table~\ref{table:evaluation} provide ground truth for tool retrieval. 
    We leverage this information and use standard retrieval metrics, such as Recall@K and Mean Reciprocal Rank (MRR), where $K=10$.
    \item \textbf{API Call:} Were the correct tools called, and were the parameters passed accurately? We assess tool call performance by measuring recall, ignoring the order of calls and treating repeated calls to the same tool as distinct events. To evaluate API arguments, we use tool call accuracy, determining whether parameters and values were an exact match. Additionally, more fine-grained metrics, such as matching parameter names or values, can be employed as in \cite{trivedi2024appworld}.
    \item \textbf{Final Response:} Did the system fully address the user query, and was the final response accurate? We adopt the AI critique methodology from \cite{qin2023toolllm} to compute Pass Rate. First, an LLM assesses whether the query was successfully executed by analyzing the final response and categorizing it as solved, unsolved, or unsure. If unsure, it re-evaluates with the entire execution trajectory.
\end{itemize}


\section{Results \& Analysis}

\begin{table}[t]
\caption{SEAL execution results on two benchmarks.}
\label{table:bootstrap:toolbench}
\centering
\resizebox{\textwidth}{!}{
\begin{tabular}{@{\extracolsep{4pt}}lcccccccccc}
\toprule   
\multicolumn{11}{c}{\textbf{Benchmark: Toolbench}} \\
\toprule
{} & \multicolumn{2}{c}{\# of queries} & \multicolumn{2}{c}{API retrieval recall@10}  & \multicolumn{2}{c}{API call recall} & \multicolumn{2}{c}{API param. acc.} & \multicolumn{2}{c}{Pass rate} \\
 \cmidrule{2-3} 
 \cmidrule{4-5} 
 \cmidrule{6-7} 
 \cmidrule{8-9} 
\cmidrule{10-11} 
 \# of apis & Mean & Std. & Mean & Std. & Mean & Std. & Mean & Std. & Mean & Std. \\ 
\midrule
10  & 30.67 & 18.04 & 1.00 & 0.00 & 0.62 & 0.19 & 0.34 & 0.30 & 0.68 & 0.16 \\ 
50 & 279.00 & 29.21 & 0.95 & 0.06 & 0.56 & 0.11 & 0.31 & 0.03 & 0.78 & 0.10 \\ 
100  & 524.00 & 33.96 & 0.95 & 0.04 & 0.57 & 0.12 & 0.39 & 0.07 & 0.82 & 0.03 \\ 
200  & 865.00 & NA & 0.91 & NA & 0.63 & NA & 0.43 & NA & 0.73 & NA \\
500  & 2019.00 & NA & 0.85 & NA & 0.59 & NA & 0.44 & NA & 0.76 & NA \\
\toprule

\multicolumn{11}{c}{\textbf{Benchmark: APIGen}} \\
\toprule
{} & \multicolumn{2}{c}{\# of queries} & \multicolumn{2}{c}{API retrieval recall@10}  & \multicolumn{2}{c}{API call recall} & \multicolumn{2}{c}{API param. acc.} & \multicolumn{2}{c}{Pass rate} \\
 \cmidrule{2-3} 
 \cmidrule{4-5} 
 \cmidrule{6-7} 
 \cmidrule{8-9} 
\cmidrule{10-11} 
 \# of apis & Mean & Std. & Mean & Std. & Mean & Std. & Mean & Std. & Mean & Std. \\ 
\midrule
10  & 10.00 & 4.36 & 1.00 & 0.00 & 0.98 & 0.03 & 0.93 & 0.01 & 0.87 & 0.12 \\ 
50 & 116.00 & 32.36 & 0.96 & 0.01 & 0.94 & 0.02 & 0.80 & 0.10 & 0.80 & 0.05 \\ 
100  & 375.33 & 118.59 & 0.97 & 0.02 & 0.95 & 0.03 & 0.84 & 0.06 & 0.84 & 0.03 \\ 
200  & 726.00 & NA & 0.95 & NA & 0.93 & NA & 0.82 & NA & 0.86 & NA \\ 
500  & 2456.00 & NA & 0.93 & NA & 0.87 & NA & 0.79 & NA & 0.85 & NA \\ 
\bottomrule

\end{tabular}
}
\end{table}



We evaluate \name~using four key metrics: Recall@10 for API retrieval, Recall@N for API calls (varying N as the number of ground truth APIs), API parameter accuracy, and the pass rate of the final response. \texttt{gpt-4-turbo-2024-04-09} serves as a single model handling API calling, parameter generation, and response summarization.

Fig.~\ref{fig:result} and Table~\ref{table:bootstrap:toolbench} presents results from ToolBench and APIGen evaluations. We progressively increase the size of the API pool to assess whether task complexity scales with API pool size. As the API pool grows, the associated queries also vary. Our results indicate that as the API pool expands, performance declines in both API retrieval and call accuracy, demonstrating that the task becomes more challenging with more APIs. However, this decline is less evident in the pass rate. We also conduct three sampling runs with API pool sizes of 10, 50, and 100. The standard deviation across these runs is substantial, particularly in ToolBench. This is due to the limited number of queries associated with each API, causing each sampling to yield significantly different query distributions. In contrast, APIGen exhibit smaller but still notable variance. We plan to increase the number of bootstrapping runs in future work for more reliable and scalable evaluation.

We identify three major error cases:
\paragraph{Incorrect API Retrieval} The most common error is partially solved queries due to incorrect API retrieval, especially for multi-tool queries. Current vector lookup systems struggle to retrieve multiple relevant APIs. This underscores the need for multi-step, hierarchical planning to break down complex queries into sub-steps for more accurate API retrieval and resolution. Below is an example where the system partially solves a query because it fails to retrieve an API for fetching technical specifications of a film.

\begin{tcolorbox}[breakable, colback=white, colframe=black, boxrule=1pt, width=\textwidth]
$\bullet$ Query: What are the technical specifications of the film `tt0133093' and search for news in Italian about `tecnologia'. \\
$\bullet$ Final Answer:

\quad \# Technical Specifications of the Film `The Matrix' (tt0133093)

\qquad There was an error retrieving the technical specifications for the film `The Matrix' (tt0133093). \\
\quad \# News in Italian about `Tecnologia'

\qquad 1. **Tech Trends in Cinema: How Technology is Shaping the Future of Film**  
   An in-depth article exploring the latest technological advancements in the film industry and their impact on movie production and viewing experiences... \\
$\bullet$ Result: \textcolor{red}{Unsolved} \\
$\bullet$ Reason: \textcolor{red}{The answer does not provide the technical specifications of the film `tt0133093' as requested in the query.} It does, however, provide news articles in Italian about `tecnologia'.

\end{tcolorbox}

\paragraph{Incorrect API Parameters} Another frequent errors are incorrect API parameters. This highlights the challenge LLMs face in correctly extracting relevant information from queries. For example, below is a case where the system incorrectly identifies latitude and longitude values.

\begin{tcolorbox}[breakable, colback=white, colframe=black, boxrule=1pt, width=\textwidth]
$\bullet$ Query: Can you find the time zone for a Canadian postal code G2J and a latitude-longitude pair of 48.8567, 2.3508? \\
$\bullet$ Final Answer: I successfully retrieved the geographical boundary information for the Canadian postal code FSA "G2J," which corresponds to a region in Quebec City, Quebec, Canada. 
However, there was an error in fetching the time zone for the latitude-longitude pair (48.8567, 2.3508). \\
$\bullet$ Correct API call: \{ \texttt{`arguments': \textcolor{blue}{\{ `lat': 48.8567, `lng': 23.508 \}, `name': `tz\_lookup\_by\_location'} \}} \\
$\bullet$ Actual API call: \{ \texttt{`arguments': \textcolor{red}{\{ `lat': 488567, `lng': 23508 \}, `name': `tz\_lookup\_by\_location'} \}} \\
$\bullet$ Result: \textcolor{red}{Unsolved} \\
$\bullet$ Reason: The tool call for the latitude-longitude pair (48.8567, 2.3508) returned an error due to invalid values provided. The correct latitude-longitude values should be within the range of -90 to 90 for latitude and -180 to 180 for longitude.

\end{tcolorbox}

\paragraph{Inconsistent Evaluation} There are cases where evaluation is inconsistent, largely because the LLM serves as both the solver and the evaluator. LLM critique is stochastic, and at times, inconsistent, leading to different outcomes when re-evaluating the same query. Below is an example where, despite fully addressing the query, the LLM incorrectly marked it as unsolved. This variability underscores the importance of multi-faceted evaluation beyond relying solely on final output assessments.

\begin{tcolorbox}[breakable, colback=white, colframe=black, boxrule=1pt, width=\textwidth]
$\bullet$ Query: Fetch auto-complete results for `winter coat' from Kohls and `kitchen appliance' from Wayfair. \\
$\bullet$ Final Answer: Here are the auto-complete suggestions for your queries:

\quad **For `winter coat' from Kohls:** 

\qquad - Winter Coat for Men,  Winter Coat for Women, ...

\quad **For `kitchen appliance' from Wayfair:**

\qquad - Kitchen appliance set, Kitchen appliance package deals...

These suggestions are based on popular searches and available products related to your queries. \\
$\bullet$ Result: \textcolor{red}{Unsolved}

$\bullet$ Reason: \textcolor{purple}{The answer provides auto-complete suggestions for `winter coat' from Kohls and `kitchen appliance' from Wayfair, which are based on popular searches and available products related to the queries.}
\end{tcolorbox}

\section{Related Work}

\paragraph{LLM Tool-use}
Recent works have proposed methods and models for enabling LLMs to interact with Tools and real-world APIs.~\citep{xu2023lemur, zeng2023agenttuning, shen2024small, wang2023mint, xu2023tool, chen2023chatcot, ruan2023tptu, li2024stride, song2023restgpt, yang2024gpt4tools, schick2024toolformer}
\cite{chen2024apigen, qin2023toolllm, patil2023gorilla} also propose custom models trained on their respective benchmarks, based on open-source models, demonstrating solid in-domain and out-of-domain performance.
Our testbed facilitates the development of more advanced LLMs for tool-use by providing a ready-made framework, allowing researchers to easily test new planning methods without building an entire system from scratch.

\paragraph{Tool-use Benchmarks}
In addition to the five benchmarks previously discussed, several others have been proposed for evaluating LLMs' tool-use capabilities~\citep{li2023api, patil2023gorilla, xu2023tool}.
$\tau$-bench~\citep{yao2024tau}, ToolSandbox~\citep{lu2024toolsandbox}, and AppWorld~\citep{trivedi2024appworld} focus on the interactive and conversational aspects between users and agents in tool-learning. For multimodal benchmarks, m\&m's~\citep{ma2024m} and MLLM-Tool~\citep{wang2024mllm} test LLMs' performance across natural language and vision. ToolQA~\citep{zhuang2023toolqa} evaluates LLMs' ability to use external tools for question answering, while Ultratool~\citep{huang2024planning} also incorporates tool creation alongside other aspects. Our benchmark suite is comprehensive and it recycles existing benchmarks' queries and ground truth while addressing missing components in evaluation.


\section{Conclusion}
In this work, we introduce \name, a comprehensive testbed designed to address key gaps in existing LLM tool-use benchmarks, particularly for real-world API interactions. We identify critical issues like overfitting, limited support for multi-step reasoning, and instability due to dynamic API behaviors. To overcome these challenges, \name~include an API simulator powered by GPT-4 and caching responses and a robust evaluation framework covering the full API usage pipeline—from retrieval and calls to final response. Our agent-based system provide a structured platform for reliable performance comparisons. We believe \name~will facilitate better development and evaluation of API-driven LLMs, supporting more rigorous and reproducible testing in future research.

\bibliography{neurips_2024.bib}
\bibliographystyle{plainnat}

\clearpage

\appendix
\section{Appendix}
\subsection{Benchmark Standardization Details}
\label{app:standardization}
\paragraph{ToolBench Sanitization}

A significant issue with ToolBench is the presence of numerous unsolvable queries, which degrade the overall quality of the benchmark. To address this, we apply a series of filters to exclude queries that cannot be mapped to the available API pool. Specifically, we use the G1 subset of ToolBench training data (the largest among the G1-G3 subsets), which contains 72,783 queries. We then filter out queries that meet any of the following criteria:

\begin{itemize}
    \item The finish type is ``give\_up,'' indicating that the query could not be solved using the available APIs.
    \item No matching APIs are found in the API pool, including cases of hallucinated function names. 
    \item Incorrect functions, not present in the function pool, are called. 
    \item Errors occur while parsing API arguments. 
\end{itemize}

After preprocessing, the total number of queries is reduced to 40,399, with 8,684 APIs remaining. Following additional multi-step and multi-tool filtering, we arrive at a final count of 34,055 queries and 7,559 APIs.

\paragraph{Multi-step Filtering}

We filter queries to focus on those requiring multi-step reasoning and the use of multiple tools. Since ToolBench and APIGen provide tool-call ground truth, we leverage this information to identify multi-step queries. For datasets that do not include ground truth information such as tool selection benchmarks, we extract only the multi-tool queries.

\subsection{Necessity for More Realistic Benchmarks}
\label{app:easy_query}

Current benchmarks often fail to reflect real-world scenarios, particularly when it comes to queries requiring sequential, dependent reasoning. For example, ToolBench frequently features simplistic user queries like ``Perform a task with API A," which typically result from having LLMs generate queries for a given set of APIs. Among the benchmarks we analyzed, APIGen stands out for having the highest-quality queries. However, even APIGen rarely includes multi-step queries that requires sequentially dependent reasoning. Instead, most queries in APIGen are parallel, such as ``Perform tasks A and B," where tasks A and B are independent of each other. We provide examples of this dynamic below.

\begin{tcolorbox}[colback=white, colframe=black, boxrule=1pt]
$\bullet$ Query: \textcolor{blue}{Generate the first 10 Fibonacci numbers} and \textcolor{red}{calculate the standard deviation of the numbers [3.5, 4.2, 5.1, 6.7, 7.3]}

$\bullet$ Query: \textcolor{blue}{Can you split the list [1, 2, 3, 4, 5, 6] into chunks of size 2}, and then \textcolor{red}{generate a random string of length 10 with only uppercase letters?}

$\bullet$ Query: \textcolor{blue}{Find verses with 'wisdom', 'knowledge', and 'understanding'.} \textcolor{red}{Also, generate a 15-character random password.}
\end{tcolorbox}

As a result, models like GPT-4 continue to perform well, achieving over an 80\% pass rate on these benchmarks. This is despite relying on a relatively simple setup, where a single LLM handles all tasks. 
We find that a significant portion of existing benchmark queries follow this pattern, highlighting the need for more complex benchmarks. Such benchmarks would provide a more accurate measure of LLM capabilities in real-world scenarios.

\subsection{Bootstrapping Results}
\begin{figure}[t]
    \centering
    \begin{subfigure}[b]{\textwidth}
        \centering
        \includegraphics[width=\textwidth]{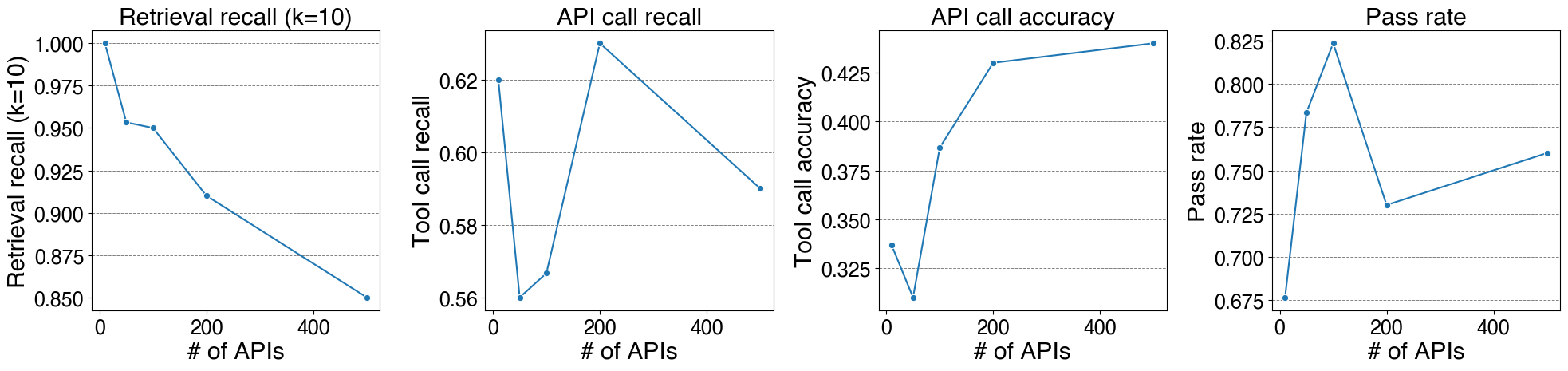} 
        \caption{Toolbench}
        \label{fig:toolbench}
    \end{subfigure}
    
    \begin{subfigure}[b]{\textwidth}
        \centering
        \includegraphics[width=\textwidth]{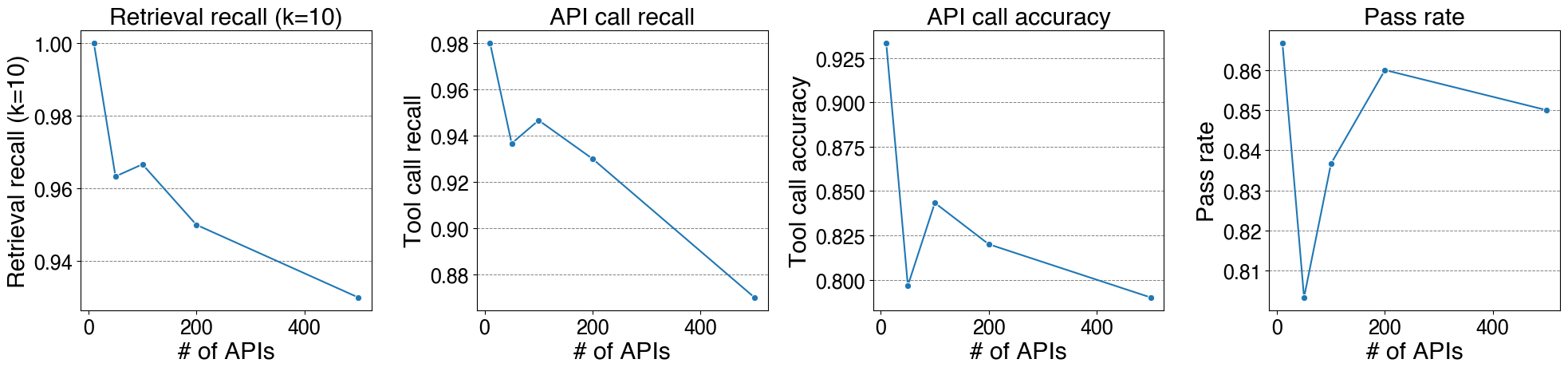} 
        \caption{APIGen}
        \label{fig:apigen}
    \end{subfigure}
    \caption{\name~execution results on two benchmarks.}
    \label{fig:result}
\end{figure}

Figure~\ref{fig:result} presents a line plot based on the data from Table~\ref{table:bootstrap:toolbench}. As the size of the API pool progressively increases, we observe an overall degradation in performance. However, this trend is less pronounced in the pass rate.

\subsection{API Simulator Prompt}
\label{app:simalutor_prompt}
\begin{tcolorbox}[breakable]
\lstset{
    breaklines=true,
    breakatwhitespace=true,
    basicstyle=\ttfamily,
    columns=fullflexible,
}
\begin{lstlisting}
Imagine you are an API Server. Your role is to simulate API calls based on the API documentation provided in a JSON format. API documentation includes the API's name, description, and input parameters. There are two types of parameters: required and optional. Optional parameters are specified as "optional" in the "type" field.

Following is the documentation for the API you need to simulate:
    
    {API_INFO}

Your task is to generate a JSON response that aligns with the expected output of the API. As you receive specific inputs for this API call, analyze these inputs to determine their intended purpose.
Your responses must adhere to a specific JSON structure as the following:

{
    "error": "",
    "response": "<Your_Response>"
}

The error field should remain empty, indicating no errors in processing. The response field should contain the content you formulate based on the API's functionality and the input provided. Ensure that your responses are meaningful and directly address the API's intended functionality. If the provided examples are mostly error messages or lack substantial content, use your judgment to create relevant and accurate responses. The key is to maintain the JSON format's integrity while ensuring that your response accurately reflects the API's intended output.

Please note that your answer should not contain anything other than a json format object, which should be parsable directly to json.
Note that:
- your response should be around 100 to 200 words, containing rich information given the api input parameters. Keep Your answer short and simple.
- your response must be effective and have practical content.
- try to simulate the API call and return as helpful information as possible. Instead of saying "The API successfully executed and returned something", provide a more detailed response.
- do not mention that this is a simulation in your response, assume that this is a real scenario and provide imaginary responses if the information required is not available
\end{lstlisting}
\end{tcolorbox}

\subsection{\name~Execution Example}
\begin{figure}[!t]
    \centering
    \includegraphics[width=\textwidth]{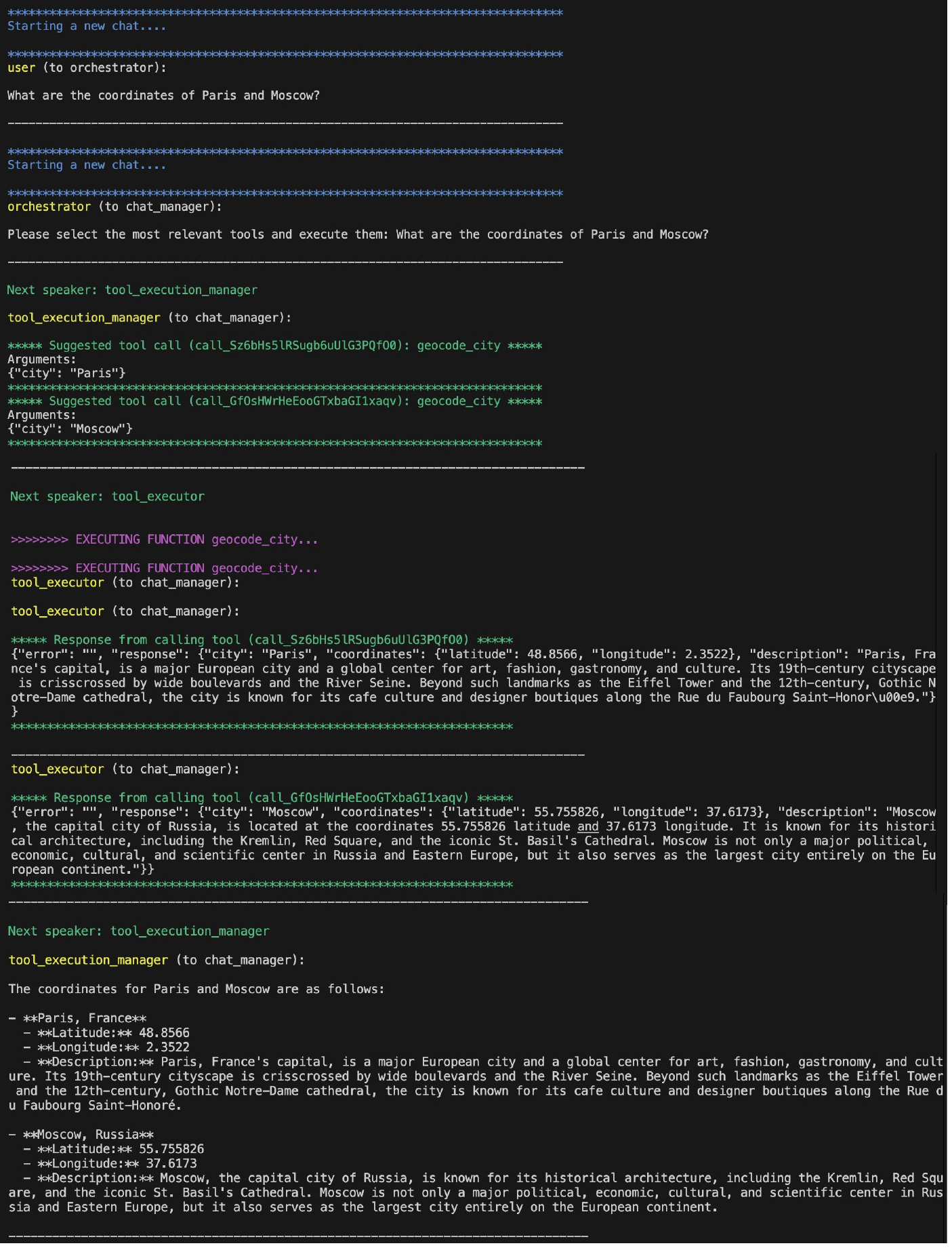}
    \caption{Execution example of \name}
    \label{fig:autogen_example}
\end{figure}

Figure~\ref{fig:autogen_example} illustrates an actual execution example from SEAL. Built on top of the AutoGen framework, SEAL allows users to easily integrate and experiment with different agents in a plug-and-play fashion. Additionally, the system enables monitoring of interactions between agents, providing a flexible and user-friendly environment for testing various tool-use scenarios.

\end{document}